# Dual-Mode Synchronization Predictive Control of Robotic Manipulator

**Zhu Dachang**[1, *], **Cui Aodong**[1], **Du Baolin**[1], **Zhu Puchen**[2]

1. School of Mechanical and Electrical Engineering, Guangzhou University, Guangzhou 510006, China
2. School of Automation, Guangdong University of Technology, Guangzhou 510006, China

* Correspondence: zdc98998@gzhu.edu.cn; Tel.: +86-020-3936-6923

**Abstract:** To reduce the contour error of the end-effector of a robotic manipulator during trajectory tracking, a dual-mode synchronization predictive control is proposed. Firstly, the dynamic model of n-DoF robotic manipulator is discretized by using the Taylor expansion method, and the mapping relationship between the joint error in the joint space and the contour error in task space is constructed. Secondly, the synchronization error and the tracking error in the joint space are defined, and the coupling error of joints is derived through the coupling coefficient $\lambda$. Thirdly, a dual-mode synchronization predictive control is proposed, and the stability of the proposed control system is guaranteed using constraint set conditions. Finally, numerical simulation and experimental results are shown to prove the effectiveness of the proposed control strategy.

## 1. Introduction

Multi-degree-of-freedom (DOF) robotic manipulator is widely used in various industrial parts because of their high accuracy and reliability. With the development of high precision machining, complex surface polishing, 3D additive machining and manufacturing, the surface quality requirements for these fields have increased significantly[1]. However, multi-DOF robotic manipulator is a typical nonlinear and strong coupling system, due to the tracking delay of single-joint and the uncoordinated dynamics of multi-degree-of-freedom system, the large contour error is caused[2]. To overcome this problem, Kuang *et al.*[3] proposed cross-coupled control, by sharing the position sensing information of each joint, the corresponding control signals are generated, and the synchronization error model is constructed to realize the synchronization of multiple motion axes[4]. Recently, a multitude of advanced contour control algorithm for multi-axis systems have been proposed, for instance, robust finite impulse response filter[5], robust trajectories following control[6], model predictive contour error control[7, 8], Iterative learning[9], on-line self tuning[10], adaptive method[11, 12], neural network control[13], sliding model control[14, 15] and so on. However, the research results demonstrate that it is extremely difficult to constitute a real-time model of contour error with trajectory tracking control of contour in task space due to the reason of contour recognition and modeling.

For real-time contour tracking control, a pre-compensating contour control strategy is proposed by Yang et al.[16] and Zhang et al.[17] which can effectively improve comprehensive performance of contour control. But the contour error in task space which defined as the minimum deviation from the actual point to the expected point, is arduous to calculate in the real-time contour tracking task[18-21]. Yeh et al.[22] used the vertical distance of the tangent from the actual point to the desired point to calculate the contour error of any trajectories, rather than the minimum deviation from the actual point to the desired point to calculate the contour error[23]. However, the contour error in task space which has been calculated should convert to joint space for correction because the manipulator is controlled in joint space.

In practice, the actual position of the robotic axial which detected by robotic axial encoder in joint space must be converted to the actual trajectory of the end-effector in task space before contour error estimation. In addition, to get contour error commands of robotic manipulator, the contour error which estimated in the task space should convert to the joint space[24]. During trajectory tracking of robotic manipulator, the reduction of single-axis tracking error in joint space does not necessarily imply that contour error of the robotic end-effector will be reduced[25]. Consequently, many researchers defined the mapping value of manipulator's task space contour error in joint space as synchronization error, and reduced the manipulator's task space contour error by reducing the synchronization error in joint space[26-28]. Because the contour error in task space has a direct mapping relationship with synchronization error, so even if certain tracking errors exist in the joint space, joint space synchronization error is zero under the condition of the contour precision of task space must also be zero. Therefore, the contour error in the task space can be effectively reduced by the synchronization error compensation of the joint space. However, most of the synchronization control algorithms of robotic are combined with PID algorithm, which is difficult to meet the requirements of rapidity and accuracy of robotic manipulator.

Model predictive control (MPC) is a modern control algorithm with low requirements on model precision, convenient online calculation, direct processing of system input and output or state constraints, and good comprehensive quality control. Compared with other manipulator control algorithms, MPC has lower requirements on model precision and better control effect. In recent years, MPC has been extensive developed in the field of robot control[29-31]. However, the stability of model predictive control of robot is arduous to prove, most of the above papers did not prove the stability of the predictive controller. According to previous discussions, a synchronous control algorithm is designed to decrease the contour error and ensure the stability of the synchronization predictive control system. The main contributions include the following:

(1) The mapping relationship of contour error between joint space and task space is established, and then, the discrete synchronous coupling error model in the joint space is constructed.

(2) A dual-mode synchronous predictive control algorithm is proposed, and the stability of the proposed control system is guaranteed by designing local control conditions and terminal constraint.

(3) Compared with computational torque controller (CTC) and traditional predictive controller in [32] and [33], numerical simulation and experimental results are shown to prove the effectiveness of the proposed control strategy.

The organization of this paper is as follows. Discretization of dynamic model of robot manipulator is derived in Section 2. Further, the design and stability analysis of the controller are presented in Section 3. Then, numerical simulation and experimental analysis are carried out in Section 4. At last, Section 5 provides a summary.

**2. Discretization of Dynamic Model of Robotic Manipulator**

The dynamic of $n-\text{DoF}$ (Degree-of-Freedom) robotic manipulators can be expressed by Newton-Euler formula as [34]

$$\tau = M(q)\ddot{q} + C(q,\dot{q})\dot{q} + G(q) \tag{1}$$

where $q \in R^{n\times 1}$ is position vector in joint space, $\tau \in R^{n\times 1}$ is generalized force vector in joint space, $M \in R^{n\times n}$ is inertia matrix, it is symmetric and positive definite, $C(q,\dot{q}) \in R^{n\times n}$ is the matrix of centrifugal and Coriolis force, $G(q) \in R^{n\times n}$ is gravity matrix.

The state variables can be defined as

$$x = \begin{bmatrix} x_1 & x_2 \end{bmatrix}^T = \begin{bmatrix} q & \dot{q} \end{bmatrix}^T \tag{2}$$

Substituting (2) into (1), the state equation of robotic manipulator can be given by

$$\begin{cases} \dot{x}_1 = x_2 \\ \dot{x}_2 = f(x_1, x_2) + P(x_1)\tau \\ y = x_1 \end{cases} \tag{3}$$

where $f(x_1, x_2) = -M(q)^{-1}(C(q,\dot{q})\dot{q} + G(q))$ and $P(x_1) = M(q)^{-1}$.

Assumed that the sampling period is $T$, and the current time is $h$, the discretization expression of $h+1$ can be obtained by using Taylor expansion method, yields to

$$x(h+1) = \begin{bmatrix} x_1(h+1) \\ x_2(h+1) \end{bmatrix} = \begin{bmatrix} x_1(h) + T\dot{x}_1(h) + \dfrac{T^2}{2}\ddot{x}_1(h) \\ x_2(h) + T\dot{x}_2(h) \end{bmatrix} \tag{4}$$

Simplifying (4) yields

$$x(h+1) = Ax(h) + B\tau(h) + G_p \tag{5}$$

where $A = \begin{bmatrix} I_n & TI_n \\ O_{n \times n} & I_n \end{bmatrix}$, $B = \begin{bmatrix} \dfrac{T^2}{2} P(x_1) \\ TP(x_1) \end{bmatrix}$, $G_p = \begin{bmatrix} \dfrac{T^2}{2} f(x_1, x_2) \\ Tf(x_1, x_2) \end{bmatrix}$.

## 3. Dual-model Synchronization Predictive Control

3.1 Mapping Relationship of Coupling Error Between Joint Space and Task Space

The desired trajectory in joint space can be defined as

$$x_r = \begin{bmatrix} x_{r1} & x_{r2} \end{bmatrix}^T \tag{6}$$

where $x_{r1}$ and $x_{r2}$ are the desired displacement and velocity vector in the joint space, respectively.

Hence, the trajectory tracking error in the joint space can be defined as

$$e = \begin{bmatrix} e_1 & e_2 \end{bmatrix}^T = x - x_r \tag{7}$$

where $e_1$ and $e_2$ are the tracking error of displacement and velocity in the joint space, respectively.

According to (5), the discretization of (7) can be given by

$$e(h+1) = x(h+1) - x_r(h+1) = Ax(h) + B\tau(h) + G_p - x_r(h+1) \tag{8}$$

The synchronization error of each axis can be calculated as follows: the contour error in the task space should be calculated firstly, and then, it will be mapped to the joint space. Finally, the mapping value of the contour error in the joint space can be adopted as the synchronization error of each axis. Furthermore, to decrease the amount of calculation, the estimated value of the closest point of the contour is used to instead of the actual value. The estimated value of the closest point of the contour is defined as the perpendicular on the tangent from the actual position to the desired position, is shown in Fig.1.

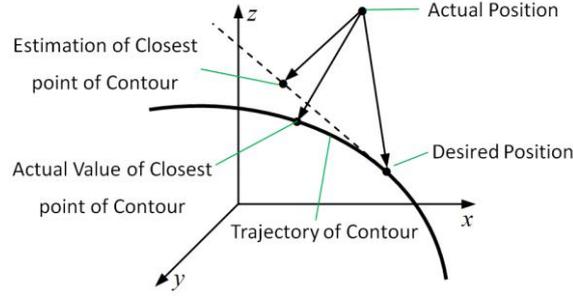

Fig.1 Estimation of contour error in the task space

Let the desired trajectory of the end-effector of robotic manipulator in the task space is $d = \begin{bmatrix} x_d(t) & y_d(t) & z_d(t) \end{bmatrix}^T$, and the contour error is $\varepsilon_o = \begin{bmatrix} \varepsilon_x & \varepsilon_y & \varepsilon_z \end{bmatrix}^T$, and the error of trajectory tracking is $e_o = \begin{bmatrix} e_x & e_y & e_z \end{bmatrix}^T$.

The contour error of the end-effector of robotic manipulator in the task space can be derived as

$$\varepsilon_o = T_c \cdot e_o \tag{9}$$

where $T_c$ is the contour error transformation matrix, and satisfied with

$$T_c = \begin{bmatrix} 1-a^2 & -ab & -ac \\ -ab & 1-b^2 & -bc \\ -ac & -bc & 1-c^2 \end{bmatrix}, \; a = \frac{\dot{x}_d(t)}{\sqrt{\dot{x}_d(t)^2 + \dot{y}_d(t)^2 + \dot{z}_d(t)^2}}, \; b = \frac{\dot{y}_d(t)}{\sqrt{\dot{x}_d(t)^2 + \dot{y}_d(t)^2 + \dot{z}_d(t)^2}}, \; c = \frac{\dot{z}_d(t)}{\sqrt{\dot{x}_d(t)^2 + \dot{y}_d(t)^2 + \dot{z}_d(t)^2}}$$

On the other hand, the kinematic Jacobian matrix of robotic manipulator mapped the position trajectory tracking error in the joint space with the task space can be given by

$$e_o = J \cdot e_1 \tag{10}$$

where $J \in R^{n \times 6}$ is kinematic Jacobian matrix of robotic manipulator with $n$-DoF.

Substituting (10) into (9), yields to

$$\varepsilon = J^{-1} \cdot \varepsilon_0 = J^{-1} \cdot T_c \cdot J \cdot e_1 \tag{11}$$

where $\varepsilon$ is defined as the synchronization error in the joint space, satisfied with $\varepsilon = J^{-1} \cdot \varepsilon_0$.

By using the coupling coefficient $\lambda$, the coupling error is defined as

$$\boldsymbol{\delta} = \boldsymbol{e} + \lambda \boldsymbol{\varepsilon} \tag{12}$$

where $\boldsymbol{\delta} \in \boldsymbol{R}^{n \times 1}$ is the coupling error of joints.

Considering the coupling between trajectory tracking error and synchronization error, the target correction trajectory is defined as $x_{rc} = \begin{bmatrix} x_{rc1} & x_{rc2} \end{bmatrix}^T$, and the coupling error can be rewritten as $\boldsymbol{\delta} = \boldsymbol{x} - \boldsymbol{x}_{rc}$ which $\boldsymbol{x}_{rc}$ can be solved by

$$\begin{cases} x_{rc1} = \dfrac{1}{1+\lambda}(x_{r1} + \lambda x_c) \\ x_{rc2} = x_{r2} \\ x_c = x - \varepsilon \end{cases} \quad (13)$$

where $x_{r1}$ and $x_{r2}$ are the position and velocity correction of joint target, respectively. $x_{rc1}$ is always between $x_{r1}$ and $x_c$ , the large the coupling coefficient is, the closer $x_{rc1}$ is to $x_c$ , the better the synchronization performance is.

Considering (8) and (12), the discretization of coupling error can be expressed as

$$\delta(h+1) = Ax(h) + B\tau(h) + G_p - x_{rc} \quad (14)$$

Substituting (12) into (14), the discretization of coupling error $\delta(h+1)$ can be given by

$$\delta(h+1) = A\delta(h) + B\tau(h) + G_p - (A+I)x_{rc} \quad (15)$$

3.2 Dual-Model Synchronous Predictive Control

The $h$-time comprehensive performance index of the robotic manipulator is defined as

$$J(h) = \sum_{i=1}^{Z}\left[\left\|\delta(h+i|h)\right\|_{Q_t}^2 + \left\|\tau(h+i-1)\right\|_{R_t}^2\right] \quad (16)$$

where $Z$ is the prediction time domain, $Q_t = \begin{bmatrix} Q_1 & \\ & Q_2 \end{bmatrix}$, and $Q_1 \in R^{n \times n}$ is position weight matrix, $Q_2 \in R^{n \times n}$ is velocity weight matrix, $R_t \in R^{n \times n}$ is the torque weight matrix.

The iteration process in the time domain is expressed as

$$X = \begin{bmatrix} x(h+1) \\ \vdots \\ x(h+Z) \end{bmatrix}, \quad U = \begin{bmatrix} \tau(h) \\ \vdots \\ \tau(h+Z-1) \end{bmatrix} \quad (17)$$

According to (5), $X$ is given by

$$X = Sx + GU + C_p \quad (18)$$

where

$$S = \begin{bmatrix} A \\ \vdots \\ A^Z \end{bmatrix}, \quad G = \begin{bmatrix} B & \cdots & 0 \\ \vdots & \ddots & \vdots \\ A^{Z-1}B & \cdots & B \end{bmatrix}, \quad C_p = \begin{bmatrix} G_p \\ AG_p + G_p \\ \vdots \\ A^{Z-1}G_p + \cdots + G_p \end{bmatrix}$$

Eq.(16) can be rewritten as

$$J = \frac{1}{2}(X - X_{rc})^T Q_t (X - X_{rc}) + \frac{1}{2}U^T R_t U \quad (19)$$

where $X_{rc} = \begin{bmatrix} x_r(h+1) \\ x_r(h+2) \\ \vdots \\ x_r(h+Z) \end{bmatrix}$.

Let $\frac{\partial J}{\partial \tau} = 0$, we obtain

$$U = -\left(G^T Q G + R\right)^{-1} G^T Q \left(Sx + C_p - X_{rc}\right) \tag{20}$$

As the rolling optimization strategy is adopted in predictive control, at $h$ time, only the first term of $U$ is applied to the robotic manipulator, and at $h+1$ time, $U$ is solved again and the first time $\tau(h+1)$ is applied to the robotic manipulator.

3.3 Stability Analysis

Three influencing factors of the stability of the dual-model synchronization predictive control system should be considered as terminal set, terminal cost function and local control law. Hence, the optimal problem of dual-model synchronization predictive control at $h$ time can be represented as

$$\min : J(h) = \sum_{i=1}^{Z-1} \left\| \delta(h+i|h) \right\|_{Q_t}^2 + \left\| \delta(h+Z|h) \right\|_P^2$$

$$s.t. \begin{cases} \delta(h+Z|h) \in \Omega = \left(\delta \in R^n \mid \delta^T P \delta \leq 1\right) \\ \delta(h|h) = \delta(h) \end{cases} \tag{21}$$

where $\left\| \delta(h+Z|h) \right\|_P^2$ is the terminal cost function, $P \in R^{n \times n}$, and $\Omega$ is the terminal set constraints.

By solving the optimal problem (21), $\delta(h+Z|h)$ can be controlled to the terminal set constraints $\Omega$ in the time domain.

Let

$$u_f(x) = \tau_f + G_P^* - B^*(A+I)x_{rc} \tag{22}$$

where $G_P^* = \begin{bmatrix} M(q) & M(q) \end{bmatrix} \begin{bmatrix} \frac{T^2}{2} & \frac{T^2}{2} \\ T & T \end{bmatrix}^{-1} G_P$, $B^* = \begin{bmatrix} M(q) & M(q) \end{bmatrix} \begin{bmatrix} \frac{T^2}{2} & \frac{T^2}{2} \\ T & T \end{bmatrix}^{-1}$, and $\tau_f$ is the actual input torque of robotic manipulator.

Combining (15) and (22), the discretization of coupling error $\delta(h+1)$ can be given by

$$\delta(h+1) = A\delta(h) + Bu_f(h) \tag{23}$$

Local control rate is given by

$$u_f(x) = H\delta(h|h) \tag{24}$$

Substituting (24) into (23), the discretization of coupling error $\delta(h+1)$ can be derived as follows

$$\delta(h+1) = (A+BH)\delta(h) \tag{25}$$

For terminal set and local control rate, the constructed $H$ and $P$ are required to make $\delta(h+Z+1|h)$ fall into terminal set $\Omega$, that is

$$\|\delta(h+Z+1|h)\|_P^2 \leq \|\delta(h+Z|h)\|_P^2 \tag{26}$$

As

$$\|\delta(h+Z+1|h)\|_P^2 - \|\delta(h+Z|h)\|_P^2 = \|(A+BH)\delta(h+Z|h)\|_P^2 - \|\delta(h+Z|h)\|_P^2 = \|\delta(h+Z|h)\|_{(A+BH)^T P(A+BH)-P}^2 \tag{27}$$

Therefore, the dual-model synchronous predictive control should be satisfied with

$$(A+BH)^T P(A+BH) - P \leq 0 \tag{28}$$

*Stability Proof*:

Assumed that the optimal performance index at $h$ time are $J^*(h)$, $U^*(h)$ and $\delta^*(h)$ which described the optimal solution and state trajectory, respectively. One of the feasible solutions of optimal problem at $h+1$ time can be given by

$$U(h+1) = \tau(h+1|h+1), \cdots, \tau(h+Z|h+1), \tau_f(h+Z+1|h+1) = \tau^*(h+1|h), \cdots, \tau_f^*(h+Z|h), \tau_f(h+Z+1|h+1) \tag{29}$$

It is noted that $U(h+1)$ is only a feasible solution and not an optimal solution, so $J(h+1)$ obtained from the feasible solution should be satisfied $J(h+1) \geq J^*(h+1)$, where $J^*(h+1)$ is the optimal performance index at $h+1$ time.

The performance index function at $h+1$ time is expressed as

$$J(h+1) = \sum_{i=1}^{Z-1} \|\delta(h+i+1|h+1)\|_{Q_t}^2 + \|\delta(h+Z+1|h+1)\|_P^2 \tag{30}$$

Considering the designed local control rate which given by (26), and $\tau(h+Z|h+1) = \tau_f^*(h+Z|h)$, we obtain

$$\delta(h+Z+1|h+1) = (A+BH)\delta(h+Z|h) \tag{31}$$

Eq.(30) can be rewritten as

$$J(h+1) = \sum_{i=1}^{Z-2} \|\delta(h+i+1|h+1)\|_{Q_t}^2 + \|\delta(h+Z|h)\|_{Q_t}^2 + \|(A+BH)\delta(h+Z|h)\|_P^2 \tag{32}$$

As

$$J^*(h) = \sum_{i=1}^{Z-1} \|\delta(h+i|h)\|_{Q_t}^2 + \|\delta(h+Z|h)\|_P^2 = \sum_{i=1}^{Z-2} \|\delta(h+i+1|h)\|_{Q_t}^2 + \|\delta(h|h)\|_{Q_t}^2 + \|\delta(h+Z|h)\|_P^2 \tag{33}$$

Combination (32) with (33), $J(h+1)$ can be given by

$$J(h+1) = J^*(h) - \|\boldsymbol{\delta}(h|h)\|_{Qt}^2 - \|\boldsymbol{\delta}(h+Z|h)\|_P^2 + \|\boldsymbol{\delta}(h+Z|h)\|_{Qt}^2 + \|(A+BH)\boldsymbol{\delta}(h+Z|h)\|_P^2$$
$$= J^*(h) - \|\boldsymbol{\delta}(h|h)\|_{Qt}^2 + \|\boldsymbol{\delta}(h+Z|h)\|_{Q+(A+BH)^T P(A+BH)-P}^2 \quad (34)$$

Assumed that $P = S^{-1}$, and $H = YS^{-1}$, (28) can be rewritten as

$$S - S^T QS - (AS + BY)^T S^{-1}(AS + BY) > 0 \quad (35)$$

The linear matrix inequality (LMI) of (35) can be expressed as

$$\begin{bmatrix} S - STQS & (AS+BY)^T \\ AS+BY & S \end{bmatrix} > 0, \quad S > 0 \quad (36)$$

The steps of the algorithm of dual-mode predictive control for robotic manipulator are given as follows

Step 1: Get the current status of the robotic manipulator.

Step 2: According to (36), $P$ and $H$ can be solved, and the optimization problem expression can be derived.

Step 3: The optimization problem (21) is solved and obtained the optimal control sequence.

Step 4: The first control rate of the optimal control sequence is applied to the robotic manipulator.

Step 5: Return to Step 1 for the next process

## 4. Simulations and Experiments

The robotic manipulator with two DOF is selected as the object in this paper, and the contour error synchronous predictive controller (CES-MPC) designed is compared with traditional predictive controller (MPC) and computational torque controller (CTC).

CTC algorithm is given by [32], and expressed as

$$\boldsymbol{\tau} = M(\boldsymbol{q})(\ddot{\boldsymbol{q}}_d + K_p \boldsymbol{e} + K_d \dot{\boldsymbol{e}}) + C(\boldsymbol{q}, \dot{\boldsymbol{q}})\dot{\boldsymbol{q}} + G(\boldsymbol{q})$$

MPC algorithm is given by [33], and expressed as

$$\boldsymbol{\tau} = -M(\boldsymbol{q})\bar{P}^{-1}\left[\frac{1}{T^2}Q_1 \boldsymbol{e}_1 + \frac{1}{T}\left(\frac{1}{2}Q_1 + Q_2\right)\boldsymbol{e}_2 + \left(\frac{1}{4}Q_1 + Q_2\right)\left(f(\boldsymbol{x}_1, \boldsymbol{x}_2) - \dot{\boldsymbol{x}}_{r2}\right)\right]$$

where $\bar{P} = \left(\frac{1}{4}Q_1 + Q_2 + T^{-4}M(\boldsymbol{q})RM(\boldsymbol{q})\right)$

The structure of robotic manipulator with two DOF is shown in Fig.2.

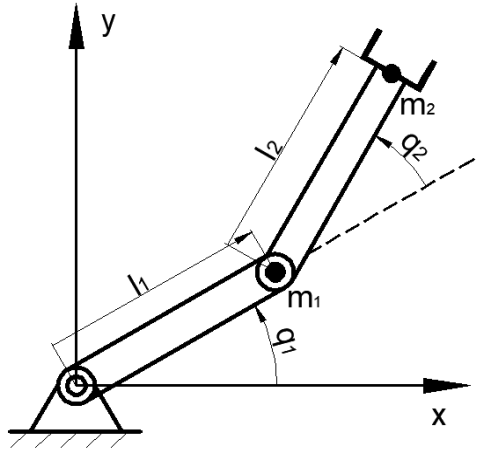

Fig.2 Structure of robotic manipulator with two DoF.

The dynamic model of the robotic manipulator with two-DoF is given by (1), and its parameters are derived as follows

$$M(q) = \begin{bmatrix} l_2^2 m_2 + 2l_1 l_2 m_2 c_2 + l_1^2 m_{12} & l_2^2 m_2 + l_1 l_2 m_2 c_2 \\ l_2^2 m_2 + l_1 l_2 m_2 c_2 & l_2^2 m_2 \end{bmatrix}, \ C(q,\dot{q}) = \begin{bmatrix} -l_1 l_2 m_2 s_2 \dot{q}_2 & -l_1 l_2 m_2 s_2 (\dot{q}_1 + \dot{q}_2) \\ l_1 l_2 m_2 s_2 \dot{q}_1 & 0 \end{bmatrix}, \ G(q) = \begin{bmatrix} l_2 m_2 g c_{12} + m_{12} l_1 g c_1 \\ l_2 m_2 g c_{12} \end{bmatrix}$$

where $c_2 = \cos(q_2)$, $c_{12} = \cos(q_1 + q_2)$, $s_2 = \sin(q_2)$, $m_{12} = m_1 + m_2$, $m_1 = 1\text{kg}$, $m_2 = 0.25\text{kg}$, $l_1 = l_1 = 0.2\text{m}$, $g = 9.8 m/s^2$.

The kinematic Jacobian matrix of the robotic manipulator with 2-DoF is calculated, yields to

$$J = \begin{bmatrix} -l_1 s_1 - l_2 s_{12} & -l_2 s_{12} \\ l_1 c_1 + l_2 c_{12} & l_2 c_{12} \end{bmatrix}$$

Selected $K_p = 300$, $K_d = 20$ in CTC, and the predictive time $N = 10$ in MPC and CES-MPC. The error weighting matrix are selected as $Q_1 = \text{diag}(10)$, $Q_2 = \text{diag}(0)$, and the torque weighting matrix is $R = \text{diag}(0)$, sampling period is 2ms, coupling coefficient is $\lambda = 1$, joint output torque constraint is $\tau_{1\max} = -\tau_{1\min} = 10\text{N} \cdot \text{m}$, $\tau_{2\max} = -\tau_{2\min} = 1\text{N} \cdot \text{m}$.

4.1 Simulations

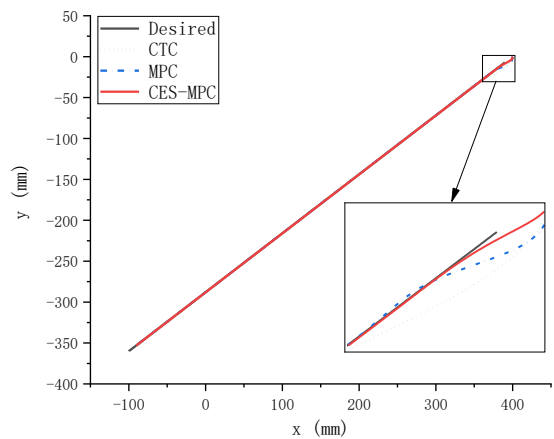

Fig.3 Trajectory tracking in the task space of robotic manipulator

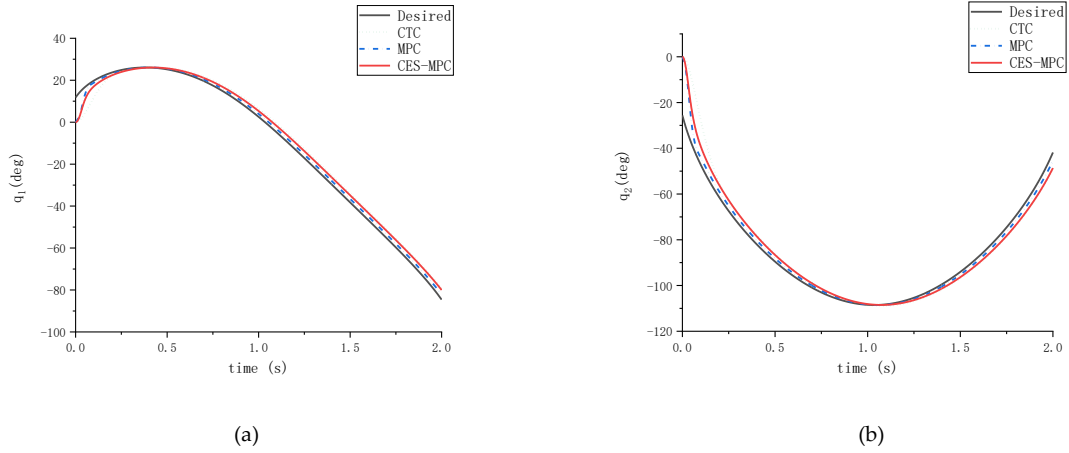

Fig.4 Trajectory tracking in the joint space of robotic manipulator. (a) joint 1#; (b) joint 2#

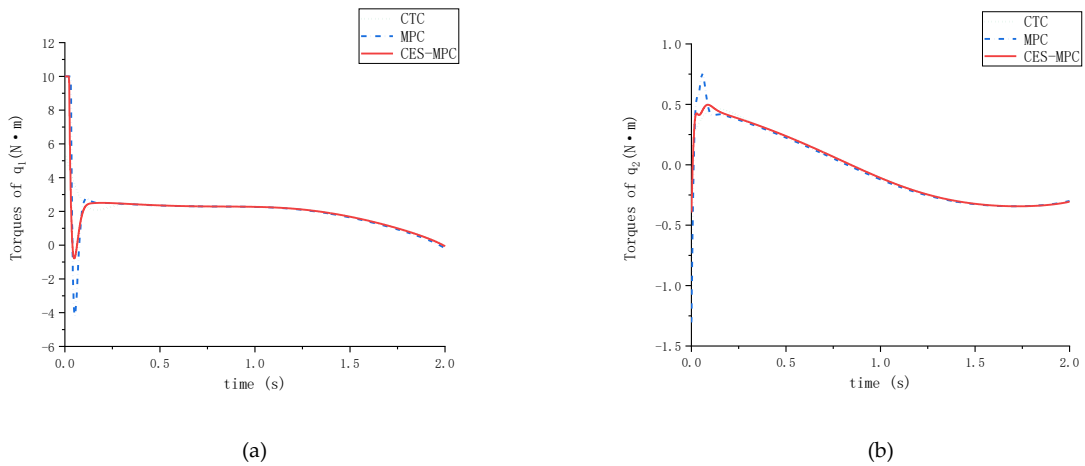

Fig.5 Torque outputs in the joint space of robotic manipulator. (a) joint 1#; (b) joint 2#

Fig.3 shows that the contour errors of the three control algorithms are relatively small without considering the friction and model error of the joint. The contour error of CES-MPC is smaller and converge faster than the algorithms of CTC and MPC. Fig.4 shows that the joint trajectory tracking error of CES-MPC is less than CTC, but delay behind MPC. It can be seen from Fig.3 that CES-MPC still has good contour accuracy when there is a large tracking error in the joint space. Fig.5 is the simulation results of torque outputs in the joint space of robotic manipulator. The simulation results show that in the initial stage of response, the torque outputs fluctuation of CES-MPC is greater than that of CTC, but less than MPC. Combined with Fig. 3 and Fig. 4, it can be seen that the contour error of CES-MPC is significantly less than CTC and MPC with small torque output fluctuation and good joint coordination.

4.2 Experiments

An experimental platform for robotic manipulator with 2-DoF is built, as shown in Fig.6.

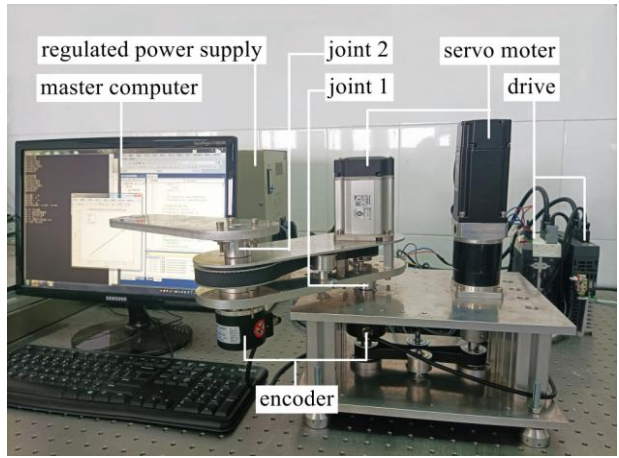

Fig.6 Experimental platform of robotic manipulator with two-DoF

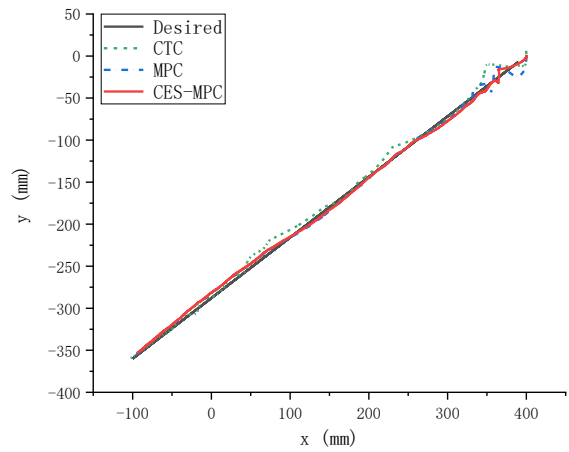

Fig.7 Trajectory tracking in the task space of robotic manipulator (with experimental platform)

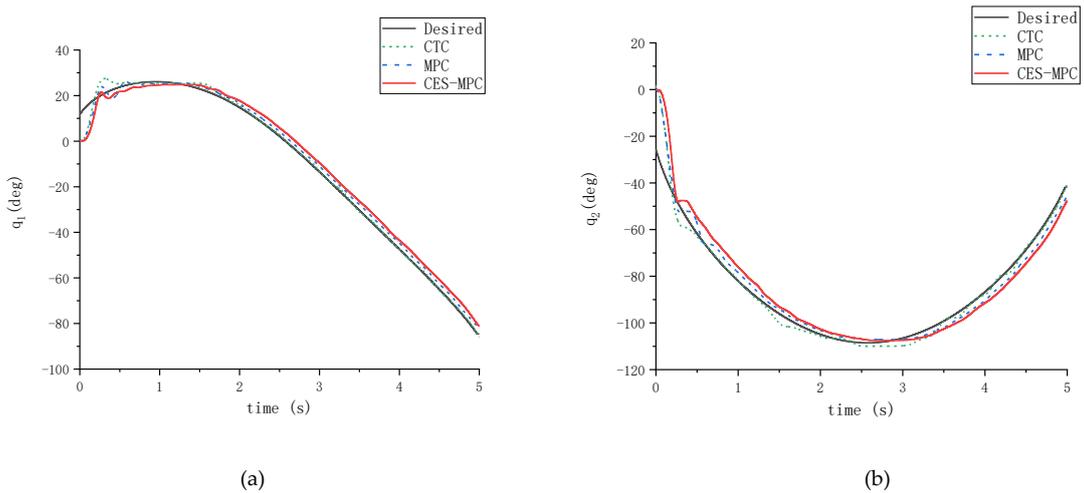

(a)                                                                                           (b)

Fig.8 Trajectory tracking in the joint space of robotic manipulator (with experimental platform). (a) joint 1#; (b) joint 2#

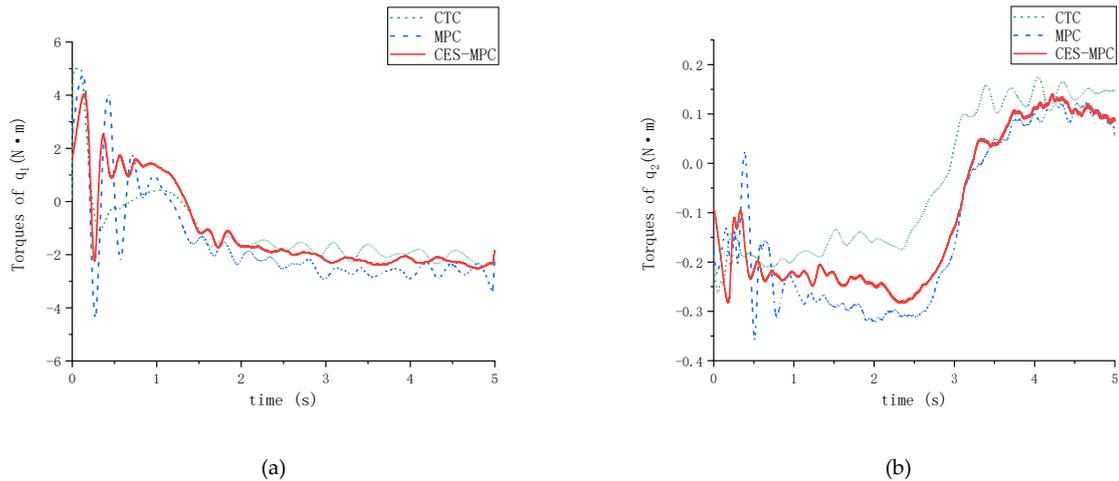

Fig.9 Torque outputs in the joint space of robotic manipulator (with experimental platform). (a) joint 1#; (b) joint 2#

Fig.7 is trajectory tracking at the end-effector of robotic manipulator, maximum contour error for CES-MPC in this part of the response is 7.4 mm, for MPC is 14.6 mm, for CTC is 21.1 mm. Experimental results demonstrate that the CES-MPC is more effective than CTC and MPC in reducing the contour error of end-effector, when the initial error exists.

Fig.8 is trajectory tracking in joint spacer of robotic manipulator, mean error for CES-MPC in this part of the response is 1.82 deg in (a) and 1.17 deg in (b), for MPC is 1.09 deg in (a) and 0.56 deg in (b), for CTC is 0.29 deg in (a) and -0.55 deg in (b). Experimental results demonstrate that the tracking error of the CES-MPC is greater than MPC and CTC, but it still has a favorable contour accuracy.

Fig.9 is torque outputs in the joint space of robotic manipulator, range of joint torque output for CES-MPC in the initial 1s in this part of the response is -2.24 to 4.03 N·m in (a) and -0.90 to -0.28 N·m in (b), for MPC is -4.37 to 4.73 N·m in (a) and -0.36 to 0.02 N·m in (b), for CTC is -0.75 to 5.00 N·m in (a) and -0.26 to -0.14 N·m in (b). The experimental results demonstrate that compared with the MPC, the torque output fluctuation of CES-MPC is smaller, and the coordination of joints is better.

By comparing the simulation and experimental results of the three algorithms, it can be seen that in the case of initial error at the end, the designed CES-MPC algorithm can more effectively reduce the end contour error than the CTC algorithm and the MPC algorithm. Moreover, compared with the conventional MPC algorithm, the joint output torque has less fluctuation and has a favorable control effect.

## 6. Conclusions

A dual-mode synchronization predictive controller for robotic manipulator is proposed, which the joint space coupling error is defined and the local controller and terminal constraint set is designed. This scheme has the following features: by reducing the joint space coupling error, the contour error of manipulator in task space during trajectory tracking is reduced. Furthermore, the local controller and terminal constraint set is designed, and the stability of the system is ensured by switched the controller to the local controller when the state of system enters the terminal constraint set. The simulation and experimental results demonstrate that the designed controller can significantly improve the contour accuracy of the robotic end-effector and make the output torque of the joint more stable in the trajectory tracking process. In the case of tracking error in joint space, it still has favorable contour accuracy of end-effector.

**Funding**

This work was supported by the National Natural Science Foundation of China (Grant numbers. 51905115), and Scientific research project of Guangzhou Education Bureau, China (Grant numbers. 202032821).

**Conflicts of Interest/Competing interests**

The authors declared that they have no conflicts of interest to this work.

**Authors' Contributions**

All authors contributed to the study conception and design. Material preparation, data collection and analysis were performed by Du Baolin, Cui Aodong and Zhu Puchen. The first draft of the manuscript was written by Zhu Dachang and all authors commented on previous versions of the manuscript. All authors read and approved the final manuscript.

**References**

1. S.Mohanty, S.P.Regalla, Y.V.D.Rao, Investigation of influence of part inclination and rotation on surface quality in robot assisted incremental sheet metal forming (RAISF), 2018, 22: 37-48.
2. Y.C. Huang, J.H. Wang, F. Wang, Event-triggered adaptive finite-time tracking control for full state constraints nonlinear systems with parameter uncertainties and given transient performance, ISA Transactions, 108(2021): 131-143.
3. Z.A.Kuang, H.J.Gao, M.Tomizuka, Precise linear-motor synchronization control via cross-coupled second-order discrete-time fractional-order sliding mode, IEEE/ASME Transactions on Mechatronics, 2021, 26(1): 358-368.
4. D.Sun, Position synchronization of multiple motion axes with adaptive coupling control, Automatica, 2003, 39(6): 997-1005.
5. B.Sencer, Y.Kakinuma, Y.Yamada, Linear interpolation of machining tool-paths with robust vibration avoidance and contouring error control, Precision Engineering, 2020, 66: 269-281.
6. Q.D.Wang, A robust trajectory similarity measure method for classical trajectory, Journal of Electronics & Information Technology, 2020, 42(8): 1999-2005.
7. Y.Gao, J.Huang, L.Chen, Constrained model predictive contour error control for feed drive systems with uncertainties, International Journal of Control Automation and Systems, 2021, 19(1): 209-220.
8. W.Kornmaneesang, S.L.Chen, MPC-based robust contouring control for a robotic machining system, 2020, 23(3): 1212-1224.
9. J.G.Li, Y.M.Wang, Y.N.Li, W.S.Luo, Reference trajectory modification based on spatial iterative learning for contour control of two-axis NC systems, IEEE/ASME Transactions on Mechatronics, 2020, 25(3): 1266-1275.
10. K.L.Li, S.Boonto, T.Nuchkrua, On-line self tuning of contouring control for high accuracy robot manipulators under various operations, International Journal of Control Automation and Systems, 2020, 18(7): 1818-1828.
11. M.Gdula, Adaptive method of 5-axis milling of sculptured surfaces elements with a curved line contour, Journal of Mechanical Science and Technology, 2019, 33(6): 2863-2872.
12. B.S.Chen, C.H.Lee, Adaptive model-free coupling controller design for multi-axis motion systems, Applied Sciences-Basel, 2020, 10(10), 3592.
13. S.X.Wang, G.Chen, Y.L.Cui, Design of contour error coupling controller based on neural network friction compensation, Mathematical Problems in Engineering, 2019, 1420380.
14. X.Du, J.Huang, L.M.Zhu, H.Ding, Sliding mode control with third-order contour error estimation for free-form contour following, Precision Engineering-Journal of the International Societies for Precision Engineering and Nanotechology, 2020, 66: 282-294.
15. X.C.Xi, W.S.Zhao, A.N.Poo, Improving CNC contouring accuracy by robust digital integral sliding mode control, International Journal of Machine Tools & Manufacture, 2015, 88: 51-61.

16. J.X.Yang, Y.Altintas, A generalized on-line estimation and control of five-axis contouring errors of CNC machine tools, International Journal of Machine Tools & Manufacture, 2015, 88: 9-23.
17. K.Zhang, A.Yuen, Y.Altintas, Pre-compensation of contour errors in five-axis CNC machine tools, International Journal of Machine Tools & Manufacture, 2013, 74: 1-11.
18. P.R.Ouyang, V.Pano, T.Dam, PID position domain control for contour tracking, International Journal of System Science, 2015, 46(1): 111-124.
19. P.R.Ouyang, V.Pano, J.Acob, Contour tracking control for multi-DOF robotic manipulators, 10th IEEE International Conference on Control and Automation (ICCA), Hangzhou, Peoples R China, Jun 12-14, 2013, 1491-1496.
20. A.Mondal, S.Ghosh, A.Ghosh, Efficient ssilhourtte-based contour tracking using local information, Soft Computing, 2016, 20(2): 785-805.
21. C.Liu, W.L.Guo, W.D.Hu, R.L.Chen, J.Liu, Real-time model-based monocular pose tracking for an asteroid by contour fitting, IEEE Transactions on Aerospace and Electronic Systems, 2021, 57(3): 1538-1561.
22. S.S.Yeh, P.L.Hsu, Estimation of the contouring error vector for the cross-coupled control design, IEEE/ASME Transactions on Mechatronics, 2002, 7(1): 44-51.
23. L.Bo, T.Y.Wang, W.Peng, Cross-coupled control based on real-time double circle contour error estimation for biaxial motion system, Measurement & Control, 2021, 54(3-4): 324-335.
24. J.Kim, M.Jin, S.H.Park, S.Y.Chung, M.J.Hwang, Task space trajectory planning for robot manipulators to follow 3-D curved contours, Electronics, 2020, 9(9), 1424.
25. X.Yang, R.Seethaler, C.P.Zhan, D.Lu, W.H.Zhao, A novel contouring error estimation method for contouring control, IEEE/ASME Transactions on Mechatronics, 2019, 24(4): 1902-1907.
26. Z.Liu, J.C.Dong, T.Y.Wang, C.Z.Ren, J.X.Guo, Real-time exact contour error calculation of NURBS tool path for contour control, International Journal of Advanced Manufacturing Technology, 2020, 108(9-10): 2803-2821.
27. X.Wu, H.Dong, J.H.She, L.Yu, High-precision contour-tracking control of Ethernet-based networked motion control systems, IEEJ Journal of Industry Applications, 2020, 9(1): 1-10.
28. D.N.Song, J.W.Ma, Y.G.Zhong, J.J.Yao, Definition and estimation of joint-space contour error based on generalized curve for five-axis contour following control, Precision Engineering-Journal of the International Societies for Precision Engineering and Nanotechology, 2020, 65: 32-43.
29. P.S.G.Cisneros, H.Werner, A velocity algorithm for nonlinear model predictive control, IEEE Transactions on Control Systems Technology, 2021, 29(3): 1310-1315.
30. E.L.Kang, H.Qiao, J.Gao, W.J.Yang, Neural network-based model predictive tracking control of an uncertain robotic manipulator with input constraints, ISA Transactions, 2021, 109: 89-101.
31. B.Brito, B.Floor, L.Ferranti, J.Alonso-Mora, Model predictive contouring control for collision avoidance in unstructured dynamic environments, IEEE Robotics and Automation Letters, 2019, 4(4): 4459-4466.
32. W.T.Huang, W.Hua, F.Y.Chen, J.G.Zhu, Enhanced model predictive torque control of fault-tolerant five-phase permanent magnet synchronous motor with harmonic restraint and voltage preselection, IEEE Transactions on Industrial Electronics, 2020, 67(8): 6259-6269.
33. S.J.Fesharaki, M.Kamali, F.Sheikholeslam, H.A.Talebi, Robust model predictive control with sliding mode for constrained non-linear systems, IET Control Theory and Apllications, 2020, 14(17): 2592-2599.
34. J.Swevers, W.Verdonck, J.De Schutter, Dynamic model identification for industrial robots-integrated experiment design and parameter estimation, IEEE Control Systems Magazine, 2007, 27(5): 58-71.